\documentclass[sigconf, superscriptaddress]{acmart}

\AtBeginDocument{%
  \providecommand\BibTeX{{%
    \normalfont B\kern-0.5em{\scshape i\kern-0.25em b}\kern-0.8em\TeX}}}

\copyrightyear{2020}
\acmYear{2020}
\setcopyright{acmcopyright}
\acmConference[SIGIR '20]{Proceedings of the 43rd International ACM SIGIR Conference on Research and Development in Information Retrieval}{July 25--30, 2020}{Virtual Event, China}
\acmBooktitle{Proceedings of the 43rd International ACM SIGIR Conference on Research and Development in Information Retrieval (SIGIR '20), July 25--30, 2020, Virtual Event, China}
\acmPrice{15.00}
\acmDOI{10.1145/3397271.3401196}
\acmISBN{978-1-4503-8016-4/20/07}

\begin{document}

\fancyhead{}

\title{Adversarial Attacks and Detection on Reinforcement Learning-Based Interactive Recommender Systems}


\author{Yuanjiang Cao}
\email{yuanjiang.cao@student.unsw.edu.au}
\affiliation{%
  \institution{University of New South Wales}
  \streetaddress{High St}
  \city{Sydney}
  \state{New South Wales}
  \postcode{NSW 2052}
}

\author{Xiaocong Chen}
\email{xiaocong.chen@student.unsw.edu.au}
\affiliation{%
  \institution{University of New South Wales}
  \streetaddress{High St}
  \city{Sydney}
  \state{New South Wales}
  \postcode{NSW 2052}
}

\author{Lina Yao}
\email{lina.yao@student.unsw.edu.au}
\affiliation{%
  \institution{University of New South Wales}
  \streetaddress{High St}
  \city{Sydney}
  \state{New South Wales}
  \postcode{NSW 2052}
}

\author{Xianzhi Wang}
\email{XIANZHI.WANG@uts.edu.au}
\affiliation{%
  \institution{University of Technology Sydney}
  \streetaddress{15 Broadway}
  \city{Sydney}
  \state{New South Wales}
  \postcode{NSW 2007}
}

\author{Wei Emma Zhang }
\email{wei.e.zhang@adelaide.edu.au }
\affiliation{%
  \institution{University of Adelaide}
  \streetaddress{North Terrace}
  \city{Adelaide}
  \state{South Australia}
  \postcode{SA 5005}
}


\begin{abstract}
  Adversarial attacks pose significant challenges for detecting adversarial attacks at an early stage.
  We propose
  attack-agnostic detection on reinforcement learning-based interactive recommendation systems. We first craft adversarial examples to show their diverse distributions
  and then augment recommendation systems by detecting potential attacks with a deep learning-based classifier based on the crafted data. Finally, we study the attack strength and frequency of adversarial examples and evaluate our model on standard datasets with multiple crafting methods. Our extensive experiments show that most adversarial attacks are effective, and both attack strength and attack frequency impact the attack performance. The strategically-timed attack achieves comparative attack performance with only 1/3 to 1/2 attack frequency. Besides, our black-box detector trained with one crafting method has the generalization ability over several crafting methods.
\end{abstract}


\maketitle


\section{Introduction}
Interactive recommendation systems capture dynamic personalized user preferences by improving their strategies continuously \cite{mahmood2007learning,thompson2004personalized, Taghipour2008hybrid}.
They have attracted enormous attention and been applied in leading companies like Amazon, Netflix, and Youtube.
The traditional methods to model user-system interactions include Multi-Armed Bandit (MAB) or Reinforcement Learning (RL). 
The former views action choices as a repeated single process, while the latter considers immediate and future rewards to model behavors' long-term benefits. RL-based systems employ a Markov Decision Process (MDP) agent that estimates the value based on both actions and states, rather than merely on actions as done by MAB.

However, reinforcement learning-based models can be fooled by small disturbances on the input data \cite{Szegedy2013Intriguing, Goodfellow2014Explaininga}.
Small imperceptible noises, such as adversarial examples, may increase prediction error or reduce reward in supervised and RL tasks---the input noise can be transferred to attack different parameters even different models, including recurrent network and RL \cite{gao2018black,Huang2017Adversarial}.
Besides, the embedding vectors of users, items and relations are piped into RL-based recommendation models, making it challenging for humans to tell the true value or to dig out the real issues in the models.
Attackers can easily leverage such characteristics to disrupt recommendation systems silently,
making defending adversarial attacks a non-trivial task for RL-based recommendation systems.

In this work, we aim to develop a general detection model to detect attacks and increase the defence ability, which provides a practical strategy to overcome the dynamic `arm-race' of attacks and to defend in the long run.
We make the following contributions:
\begin{itemize}
    \item We systematically investigate adversarial attacks and detection approaches with a focus on reinforcement learning-based recommendation systems and demonstrate the effectiveness of the designed adversarial examples and strategically-timed attack. 
   \item We propose an encoder-classification detection model for attack-agnostic detection. The encoder captures the temporal relationship among sequence actions in reinforcement learning. We further use an attention-based classifier to highlight the critical time steps out of ample interactive space.
   \item We empirically show that even small perturbations can reduce the performance of most attack methods significantly.
   Our statistical validation shows that multiple attack methods generate similar actions of the attacked system, providing insights into improving the detection performance. 
\end{itemize}

\begin{figure}[!ht]
  \centering
    \includegraphics[width=0.5\textwidth]{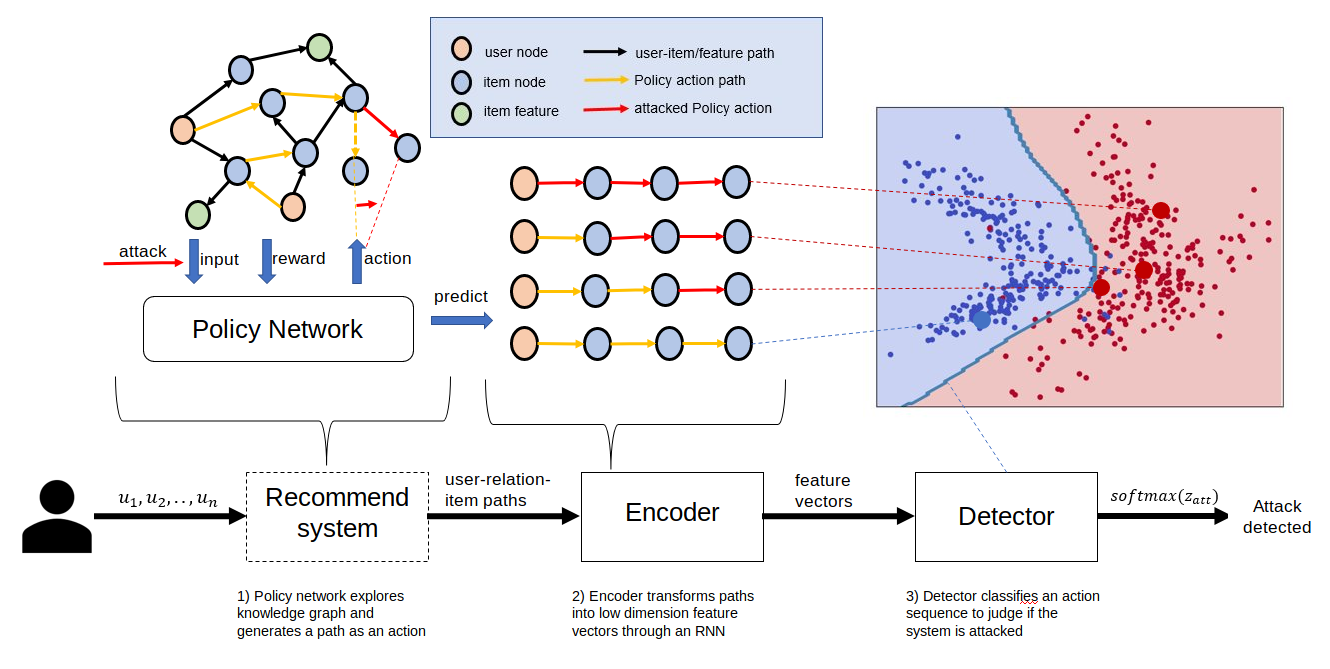}
  \centering
  \caption{
  Our proposed Adversarial Attack and Detection Approach for RL-based Recommender Systems.
  }
  \label{fig-attention-0}
\end{figure}

\section{Methodology}
\label{method}


\subsection{RL-based Interactive Recommendation}
\label{problem-definition}
Interactive recommendation systems suggest items to users and receives feedback. Given a user $u_{j} \in U= \{u_0, u_1, u_2, ..., u_n\}$, a set of items $I = \{i_0, i_1, i_2, ..., i_n\}$, and the user feedback history $i_{k_1}, i_{k_2}, ..., i_{k_{t-1}}$, the recommendation system suggests a new item $i_{k_{t}}$.
This problem represents a Markov Decision Process as follows:
\begin{itemize}
    \item State ($s_t$): a historical interaction between a user and the recommendation system computed by an embedding or encoder module.
    \item Action ($a_t$): an item or a set of items recommended by the RL agent.
    \item Reward ($r_t$): a variable related to a user's feedback to guide the reinforcement model towards true user preference. 
    \item Policy ($\pi(a_t | s_t)$): a conditional probability distribution of items which the agent might recommend to a user $u_i$ given the state of last time step $s_{t-1}$. 
    The learning process aims to get an optimal policy.
    \item Value function ($Q(s_t, a_t)$): the agent's estimation of reward of current states $s_t$ and recommended item $a_t$. We define the reward as the cosine similarity between user and item embedding vectors.
\end{itemize}
The reinforcement agent could follow REINFORCE with baseline or Actor-Critic algorithm that both consist of a value network and a policy network \cite{xian2019reinforcement}. The attack model may generate adversarial examples using either the value network \cite{Huang2017Adversarial} or the policy network\cite{Pattanaik2017Robust}.

\subsection{Attack Model}
\label{method-attack}

\vspace{1mm}\noindent\textbf{FGSM-based attack.} We define an adversarial attack as an additional small perturbation $\delta$ on benign examples $x$, which can be a composition of embedding vectors of users, relations and items \cite{xian2019reinforcement}.
Unlike perturbations on images or texts, $\delta$ can be large due to the enormous manual work to check the embedding vectors. We define an adversarial example as
\begin{equation}
\begin{aligned}
    & \min\limits_{\delta} R_T = \sum\limits_{t=1}\limits^{T} Q(s_t + \delta,\ a_t).\\
    & a_t = \pi^{*}(a_t | s_t + \delta) \quad  \text{subject to}\  S(s_t, s_t+\delta) \leq l
\end{aligned}
\end{equation}

\vspace{1mm}\noindent\textbf{Attack with smaller frequency.} The strategically-timed attack \cite{Lin2017Tacticsc} aims to decreases the attack frequency without sacrificing the performance of the un-targeted reinforcement attack.
We formally present it below:
\begin{equation}
\begin{aligned}
    & \delta_t = \delta_t * c_t \quad c_t \in \{0, 1\},\quad \frac{\sum_{t=1}^{T}c_t}{T} < d\\
\end{aligned}
\end{equation}
where $c_t$ is a binary variable that controls when to attack; $d < T$ is the frequency of adversarial examples. There are two approaches to generate the binary sequence $c_{1:T}$ optimizing a hard integer programming problem and generating sequences via heuristic methods. Let $p_0, p_1$ be the two maximum probability of an policy $\pi$, $c_t$ be the attack mask on time step, which is different from \cite{Lin2017Tacticsc}:
$$
c_t = (p_0 - p_1) > threshold
$$
We pick out actions that have the max distance between the highest two probabilities, which means we attack on the most confident actions of the agent. Experiments show that this strategy works.
In contrast, Jacobian-based Saliency Map Attack (JSMA) \cite{papernot2016limitations} and Deepfool \cite{moosavi2016deepfool} are based on the gradient of actions rather than the gradient of $Q$ value. One key component of JSMA is saliency map computation, which decides which dimension of vectors (in Image classification is pixels) are modified. Deepfool pinpoints the attack dimension by comparison of affine distances between some class and temporal class.

\subsection{Detection Model} 

The detection model is a supervised classifier detects adversarial examples with actions of the reinforcement agent.
Suppose the action distributions of an agent are shifted by adversarial examples (Section \ref{exp} shows statistical evidence of the drift).
Given an abnormal action sequence $a = \pi^{*}(a|s + \delta)$, the detection model aims to establish separating hyperplane between adversarial examples and normal examples, thereby measuring the probability $p(y | a, \theta)$ or $p(y | \pi^{*}, s, \delta, \theta)$, where $y$ is a binary variable indicating whether the input data are attacked.

To detect the adversarial examples presented in the last section, we employ an attention-based classifier.
The detection model consists of two parts. The first is a GRU encoder, to encode the action methods into a low dimensional feature vector. The second is an attention based decoder with classifier to separate different data. This encoder-decoder model has a bottleneck structure that filters out noisy information. The formulation of GRU is as follows:
\begin{equation}
\begin{aligned}
    & z_t = \sigma_g(W_z a_t + U_z h_{t-1})\\
    & r_t = \sigma_g(W_r a_t + U_r h_{t-1})\\
    & \hat{h_t} = tanh(W_h a_t + U_h \circ h_{t-1})\\
    & h_t = (1 - z_t) \circ h_{t-1} + z_t \circ \hat{h_t}
\end{aligned}
\end{equation}

We use $a_{1:T}$ to denote action sequence, which is a series of user relation vectors or item embedding vectors. $h_{t}$ is the output of encoder.

The attention-based decoder is formulated below.
\begin{equation}
\begin{aligned}
    & \alpha_{t} = Softmax(W_{e} e + b_{e})_{t}\\
    & out_{t+1}, hid_{t+1} = GRU(\alpha_{t} a_t, hid_{t}) \\
    & p_t = Sigmoid(W_{out} out_t + b_{out})
\end{aligned}
\end{equation}
where $e$ is the combined vector of action embedding, hidden states $hid$ and encoder output, the combination method is a linear unit with ReLU activation. GRU is reused to generate multiple hidden states at each time steps. Loss is computed at each time step.
After processed through the attention model and GRU, the vector is then piped into a linear unit with sigmoid function to predict if the agent is polluted. The loss function is the cross entropy between the true label and corresponding probability,
$$
J(Att(a_{1:T}), y) = - y \circ log(p)
$$

\section{Experiments}
\label{exp}
In this section, we report our experiments to evaluate attack methods and our detection model.


\subsection{Dataset and Experiment Setup}
\label{exp-dataset}
We conduct experiments following \cite{chen2019large} and \cite{xian2019reinforcement} over a \textit{Amazon dataset} \cite{he2016ups}. This public dataset contains user reviews and metadata of the Amazon e-commerce platform from 1996 to 2014. We utilize three subsets, namely Beauty, Cellphones, and Clothing, which are originally provided by \cite{xian2019reinforcement} on Github . Details about Amazon dataset analysis can be found in \cite{xian2019reinforcement}.


Our experiments are based on \cite{xian2019reinforcement}.
During dataset preprocessing,feature words with higher TF-IDF scores than 0.1 are filtered out.
70\% of data in each dataset comprises the training set (the rest are the test set). We take actions of reinforcement agent as the detection data.
We define the actions of PGPR \cite{xian2019reinforcement} as heterogeneous graph paths that start from users and have a length of 4.
The three Amazon sub-datasets (Beauty, Cellphones, and Clothing) contain 22,363, 27,879, and 39387 users, respectively.
To accelerate experiments, we use the first 10,000 users of each dataset to produce adversarial examples. Users in Beauty get, on average, 127.51 paths. The counterparts for Cellphones and Clothing are 121.92 and 122.71. As the number of paths is large, we utilize the first 100,000 paths for train and validation with split ratio 80/20.  We randomly sampled 100,000 paths from each action file to form the test set. 

We attack trained RL agent with methods in section \ref{method-attack}. We slightly modify JSMA and Deepfool for our experiments---we create the saliency map by calculating the product of the target label and temporal label and achieve both effectiveness and higher efficiency (by 0.32 seconds per iteration) of JSMA;
we decrease the computation load on a group of gradients on Deepfool by sampling.
Besides, we set the hidden size of the GRU to 32 for the encoder, the drop rate of the attention-based classifier to 0.5, the maximum length of a user-item path to 4, the learning rate and weight decay of the optimization solver, Adam, to 5e-4 and 0.01, respectively.

\subsection{Attack Experiments}

\vspace{1mm}\noindent\textbf{Adversarial attack results.} We are interested in how vulnerable the agent is to perturbation in semantic embedding space. An attack will be effective if a small perturbation leads to a notable performance reduction. 
We reuse the evaluation metrics of \cite{xian2019reinforcement}, namely Normalized Discounted Cumulative Gain (NDCG), Recall, Hit Ratio (HR), and Precision for evaluation. All metrics are computed based on the top 10 sorted predictions for each user, and they are presented in percentage without specific notion.

Table \ref{tab-basic-attack} shows the performance of different attack methods. Attack results share the same trend with the distribution discrepancy. Most attack methods significantly reduce the performance of the reinforcement system. FGSM $l_1$ achieves the best performance. It reveals that single dimension attack can change the agent's action drastically. While FGSM $l_2$ is less effective, where the metrics just fluctuates around the original baseline (Table \ref{tab-basic-attack}), partly because of small disturbance created by $l_2$ method. Specifically, JSMA achieves comparable results as $l_{inf}$ with a small attack region. 
Attacks on Clothing and Cellphones sub-datasets show similar effects.

\begin{table*}
  \caption{Adversarial attack results, MMD between benign distribution and adversarial distribution on Amazon Beauty}
  \centering
  \label{tab-basic-attack}
    \begin{tabular}{c c c c c c|| c c c}
    \toprule
    Data &Parameters &NDCG &Recall &HR &Precision &MMD-org &MMD-$l_1$\\
    \hline
    Original &- &4.654 &6.572 &13.993 &1.675 &0.121 &0.620\\
    FGSM $l_1$ &$\epsilon=$0.1 &2.695 &3.714 &6.599 &0.693 &0.604 &0.010\\
    FGSM $l_2$ &$\epsilon=$1.0 &4.567 &6.555 &13.751 &1.653 &0.016 &0.573\\
    FGSM $l_{inf}$ &$\epsilon=$0.5 &2.830 &3.909 &7.351 &0.787 &0.570 &0.011 \\
    JSMA &- &2.984 &3.844 &8.254 &0.931 &0.412 &0.034\\
    Deepfool &- &3.280 &4.352 &9.548 &1.050 &0.177 &0.458\\
    \bottomrule
\end{tabular}
\end{table*}

 

\begin{figure}[!htb]
  \centering
    \includegraphics[width=0.5\textwidth]{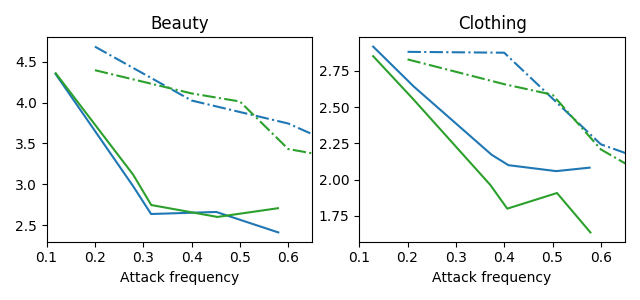}
  \centering
  \caption{NDCG of attack frequency on Beauty and Clothing subsets. Dashdot lines represent random attacks, solid lines are strategically-timed attacks. Blue and green lines are FGSM $l_{inf}$ and  $l_1$ attacks respectively.}
  \label{fig-attack-time}
\end{figure}

\vspace{1mm}\noindent\textbf{Impact of attack frequency.}
We conduct two experiments on attack frequency, random attack and strategically attack. The difference is if adversarial examples are crafted with a frequency parameter $p_{freq}$, or generated by the method shown in Section \ref{method-attack}. The NDCG metric is presented in Figure \ref{fig-attack-time}; other metrics have a similar trend. It can be seen from \ref{fig-attack-time} that the random attack performs worse than the strategically-timed attack. With strategically timed method, attacking $\frac{1}{3} - \frac{1}{2}$ time steps achieves a significant reduction in all metrics.

\vspace{1mm}\noindent\textbf{Analysis of adversarial examples.}
We use Maximum Mean Discrepancy as statistical measures to capture distribution distance.
This divergence is defined as:
$$
MMD(k, X_{org}, X_{adv}) = \sup\limits_{k \in K}\Bigg(\frac{1}{n}\sum\limits_{i=1}^{n}k(x_{org,i}) - \frac{1}{m}\sum\limits_{i=1}^{n}k(x_{adv,i})\Bigg)
$$
where $k$ is the kernel function, i.e., a radial basis function,
$X_{org}, X_{adv}$ are benign and adversarial examples. 

MMD-org reveals the discrepancy between the original and adversarial datasets. And MMD-$l_1$ presents the discrepancy among different attack methods. The results (Table \ref{tab-basic-attack}) show that the adversarial distribution is different from the original distribution. Also, the disturbed distributions are closed to each other regardless of the attack type.
This insight makes it clear that we can use a classifier to separate benign data and adversarial data and it can detect several attacks at the same time, which might be transferred to other reinforcement learning attack detection tasks.

\subsection{Detection Experiments}


From a statistical perspective, the above analysis shows that one classifier can detect multiple types of attacks.
We evaluate the detection performance using Precision, Recall and F1 score.

Our attention-based detection model is trained on FGSM $l_1$ attack with $\epsilon$ at 0.1 and detects all attack types. The results (Table \ref{tab-detection-basic}) show that the attack are stronger, the model achieves better performance.
$l_{\infty}$ attack validates this trend, which shows that our model can detect weak attacks as well. The result of detection on $l_2$ attack can be reasoned with MMD analysis shown above, high precision and low recall show that most $l_2$ adversarial examples are close to benign data which confuses the detector.
The $l_1$ attack with $\epsilon=1.0$ validates that our detector performs well yet achieves worse performance on other tests of Cellphones dataset.

\begin{table}[!htb]
  \caption{Detection Result \& Factor Analysis}
  \centering
  \label{tab-detection-basic}
  \begin{tabular}{cccccc}
    \toprule
    Dataset &Attack &Precision &Recall &F1 Score\\
    \hline
    Beauty &$l_1$ 0.1 &0.919 &0.890 &0.904\\
           &$l_2$ 1.0 &0.605 &0.119 &0.199\\
           &$l_{inf}$ 0.5 &0.918 &0.871 &0.894\\
           &JSMA &0.910 &0.793 &0.848\\
           &Deepfool &0.915 &0.840 &0.876\\
    \hline
    Cellphones &$l_1$ 0.1 &0.801 &0.781 &0.791\\
           &$l_2$ 1.0 &0.754 &0.593 &0.664\\
           &$l_{inf}$ 0.5 &0.795 &0.752 &0.773\\
           &$l_1$ 1.0 &0.810 &0.825 &0.817\\
    \hline
    Clothing &$l_1$ 0.1 &0.911 &0.866 &0.888\\
           &$l_2$ 1.0 &0.541 &0.099 &0.168\\
           &$l_{inf}$ 0.5 &0.912 &0.879 &0.895\\
    \midrule
    Dataset &Frequency &Precision &Recall &F1 Score\\
    \hline
    Beauty &$l_1$ 0.02 &0.823 &0.362 &0.503\\
           &$l_1$ 0.08 &0.918 &0.872 &0.894\\
           &$l_1$ 0.3 &0.922 &0.927 &0.924\\
    \midrule
    Dataset &Frequency &Precision &Recall &F1 Score\\
    \hline
    Beauty &$l_1$ 0.579 &0.921 &0.912 &0.917\\
           &$l_1$ 0.316 &0.918 &0.879 &0.898\\
           &$l_1$ 0.118 &0.837 &0.401 &0.543\\
    \bottomrule
\end{tabular}
\end{table}

Our results on factor analysis (Table \ref{tab-detection-basic}) show that the detection model can detect attacks even under low attack frequencies.
But the detection accuracy decreases as the attack frequency drops---the recall reduces significantly to 40.1\% when 11.8\% examples represent attacks.


\section{Conclusion}
Adversarial attacks on reinforcement learning-based recommendation system can degrade user experience.
In this paper, we systematically study adversarial attacks and their factor impacts. We conduct statistical analysis to show classifiers, especially an attention-based detector, can well separate the detection data. Our extensive experiments show both our attack and detection models achieve satisfactory performance.

\bibliographystyle{ACM-Reference-Format}
\bibliography{ref}

\end{document}